\documentclass[10pt, journal]{IEEEtran}
\usepackage{cite}
\usepackage[dvips]{graphicx}
\usepackage{url}
\usepackage{fancyhdr}

\begin{document}

\pagestyle{fancy}
\lhead{A pre-print version; final version published in D. Dietrich et al. (Eds.): Simulating the Mind. Springer 2009.\\*
http://www.springer.com/springerwiennewyork/computer+science/book/978-3-211-09450-1}

\title{\LARGE \bf A computational model of affects}

\author{Mika Turkia\\
  University of Helsinki\\
  \texttt{turkia@cs.helsinki.fi}
  \thanks{Manuscript received January 31, 2008; revised March 15, 2008.}
}
  
\maketitle

\begin{abstract}
Emotions and feelings (i.e. affects) are a central feature of human behavior. 
Due to complexity and interdisciplinarity of affective phenomena, attempts to define them have often been unsatisfactory. 
This article provides a simple logical structure, in which affective concepts can be defined. 
The set of affects defined is similar to the set of emotions covered in the OCC model \cite{occ88}, but the model presented in this article is fully computationally defined, whereas the OCC model depends on undefined concepts. 

Following Matthis \cite{matthis00}, affects are seen as unconscious, emotions as preconscious and feelings as conscious. Affects are thus a superclass of emotions and feelings with regards to consciousness. A set of affective states and related affect-specific behaviors and strategies can be defined with unconscious affects only. 

In addition, affects are defined as processes of change in the body state, that have specific triggers. For example, an affect of hope is defined as a specific body state that is triggered when the agent is becomes informed about a future event, that is positive with regards to the agent's needs. 

Affects are differentiated from each other by types of causing events. 
Affects caused by unexpected positive, neutral and negative events are \emph{delight}, \emph{surprise} and \emph{fright}, respectively. Affects caused by expected positive and negative future events are \emph{hope} and \emph{fear}. 

Affects caused by expected past events are as follows: \emph{satisfaction} results from a positive expectation being fulfilled, \emph{disappointment} results from a positive expectation not being fulfilled, \emph{fears-confirmed} results from a negative expectation being fulfilled, and \emph{relief} results from a negative expectation not being fulfilled. Pride is targeted towards a self-originated positive event, and shame towards a self-originated negative event. Remorse is targeted towards a self-originated action causing a negative event. Pity is targeted towards a liked agent experiencing a negative event, and happy-for towards a liked agent experiencing a positive event. Resentment is targeted towards a disliked agent experiencing a positive event, and gloating towards a disliked agent experiencing a negative event. An agent is liked/loved if it has produced a net utility greater than zero, and disliked/hated if the net utility is lower than zero. An agent is desired if it is expected to produce a positive net utility in the future, and disliked if the expected net utility is negative. 

The above model for unconscious affects is easily computationally implementable, and may be used as a starting point in building believable simulation models of human behavior. The models can be used as a starting point in the development of computational psychological, psychiatric, sociological and criminological theories, or in e.g. computer games. 
\end{abstract}

\thispagestyle{fancy} 

\section{Introduction}

\PARstart{I}{n} this article, computationally trivial differentiation criteria for the most common affects for simple agents are presented (for introduction to the agent-based approach see e.g. \cite{russell95}). The focus is in providing a simple logical or computational structure, in which affective concepts can be defined. 

The set of affects defined is similar to the set of emotions presented in the OCC model \cite{occ88}, which has been a popular emotion model in computer science. However, as e.g. Petta points out, it is only partially computationally defined \cite{petta04}. For example, definitions of many emotions are based on concepts of standards and norms, but these concepts are undefined. 
These limitations have often not been taken into account. 

The OCC model may be closer to a requirements specification than to a directly implementable model. Ortony himself has later described the model as too complicated and proposed a simpler model \cite{ortony03}, which may however be somewhat limited. 

The missing concepts are however definable. In this article, the necessary definitions and a restructured model similar to the OCC model are presented. A simple implementation of the structural classification model is also presented. 

The primary concept is the concept of a computational agent, that represents the affective subject. An agent is defined as possessing a predefined set of goals, e.g. self-survival and reproduction. 
These goals form the basis of subjectively experienced utility. An event fulfilling a goal has a positive utility; correspondingly, an event reducing the fulfillment of a goal has a negative utility. 
All other goals may be seen as subgoals derived from these primary, evolutionarily formed goals. 
Utility is thus seen as a measure of evolutionary fitness. 

An agent is defined as logically consisting of a controlling part (nervous system) and a controlled part (the rest of the body). To be able to control its environment (through controlling its own body, that performs actions, that affect the environment) the agent forms a model of the environment. This object model consists of representations of previously perceived objects associated with the utilities they have produced. All future predictions are thus based solely on past experiences. 

An affect is defined as a process, in which the controlling part, on perceiving an utility-changing event in the context of its current object representations (object model), produces a corresponding evolutionarily determined bodily change, i.e. transfers the agent to another body state. 

Specific behaviors and strategies can be associated with specific affective states. 
The set of possible affective states and associated actions may be predefined (i.e. innate) or learned. Innate associations may include e.g. aggression towards the causing object of a frustrating event (i.e. aggression as an action associated with frustrated state). Learned actions are acquired by associating previously experienced states with the results of experimented actions in these states. 

Emotions and feelings are defined as subclasses of affects \cite{matthis00}. 
Emotions are defined as preconscious affects and feelings as conscious affects. 
Being conscious of some object is preliminarily defined as the object being a target of attention (see e.g. \cite{baars97}). Correspondingly, being preconscious is being capable of attending to the object when needed. In contrast, unconscious affects are processes that cannot be perceived at all due to lack of sensory mechanisms, or otherwise cannot be attended to, due to e.g. limitations or mechanisms of the controlling part. 

Thus, emotions and feelings are conceptualized as requiring the agent to be capable of being conscious of changes in its body states \cite{damasio03spinoza}. 
As an affect was defined as a physiological state change triggered by a perception of a predefined event or an object constellation, we can also define a system where agents are not conscious of their affects, but still have them. These unconscious affects suffice to produce a set of states, to which affect-specific behaviors and strategies may be bound. In effect, such agents are affective but not emotional. 

Relations between the concepts of affect, emotion, feeling and consciousness were defined above. 
Another question is the differentiation of affects from each other. This is achieved by classifying the triggering object constellations. The constellations include the state of the agent, which in turn includes the complete history of the agent. 
In other words the idea is the following: affects are differentiated from each other by both the event type and the contents of the current object model, i.e. by the structure of the subjective social situation. This subjective social situation is formed by agent's life history, i.e. the series of all perceived events. 

To preliminarily bind this conceptual framework to psychoanalytic object relations theory (e.g. \cite{fonagy03,tahka93}), we note that objects and their utilities in relation to self form a network of \emph{object relations}.

\section{Simulation environment}
\label{agents}

\subsection{Ontological definitions}

Let us assume a \emph{world} that produces a series of \emph{events}. 
The world contains \emph{objects}, some of which are alive. Living objects that are able to act on the world are called \emph{agents}. Agents' \emph{actions} are a subset of events. 
An event consists of a type indicator and references to causing object(s) and a target object(s). 

\subsection{An agent as a control system}

An agent is seen as a \emph{control system}, which consists of two parts: a controlled system and a controlling system (in computer science, this idea has been presented by at least \cite{sloman93}). This division can be done on a functional or logical level only. Let us thus define, that an agent's controlling system is the brain and the associated neural system, and the controlled system is the body. Physically the controlling system is part of the controlled system, but on a functional level they are separate, although there may be feedback loops, so that the actions of the controlling system change its own physical basis, which in turn results in modifications in the rules of control. 

An agent usually experiences only a part of the series of events in the world; that part is the \emph{environment} of the agent. Experiencing happens through \emph{perceptions} that contain events.  These events are \emph{evaluated} with regards to target agent's needs. The value of an event for an agent's needs can be called its \emph{utility} (the utility concept used here is similar to the utility concept used in reinforcement learning \cite{sutton98}). The utility of an event or action is associated with the causing object.  

In a simplified model, fixed utilities can be assigned to event types. In this case the evaluation phase is omitted.  

The basis of utilities represented in the controlling system (i.e. mind) are the needs of the controlled system (i.e. body). Utilities direct actions to attempt the fulfillment of the needs of the body. Utility is thus to be understood as a measure of change in evolutionary fitness caused by an event. An agent attempts to experience as much value as possible (maximize its utility) during the rest of its lifetime. Maximizing utility maximizes evolutionary fitness, i.e. self-survival and reproduction, according to an utility function preset by evolution of the species in question. 

In order to attempt this utility maximization, the agent has to be able to affect the environment (to act), so that it can pursue highly valued events and try to avoid less valued events. 
It also has to be able to predict which actions would lead it to experience highly valued events and avoid low-valued (meaningless or harmful) experiences. 
To be able to predict, the agent has to have \emph{a model of the environment}, which contains models of perceived objects associated with their utilities. 

As the agent can never know if it has seen all possible objects and event types of the world, all information is necessarily uncertain, i.e. probability-based in its nature. Therefore, when an agent performs an action, it expects a certain utility. The actual resulting utility may differ from expected, due to the necessarily limited predictive capability of the internal model of the environment. 

At any moment, an agent selects and performs the action with the highest expected utility. Thus, every action maximizes subjective utility. Also, any goal is derived from the primary goals (needs), i.e. self-survival and reproduction. 

\emph{Lifetime utility} means the sum of all value inputs that the agent experiences during its lifetime.
\emph{Past utility} means the sum of already experienced value inputs. 
\emph{Future utility} means the sum of value inputs to be experienced during the rest of the lifetime. Since this is unknown, it can only be estimated based on past utility. 
This estimate is called \emph{expected future utility}.
Then \emph{expected lifetime utility} is the sum of past utility and expected future utility.
An agent maximizes expected future utility. If it would maximize expected lifetime utility, it would die when the expected future utility falls below zero. 

\subsection{Temperament and personality}

\emph{Personality} is the consequences of learned differences expressed in behavior. Thus, personality is determined by the learned contents of the controlling system. 
\emph{Temperament} is the consequences of physiological differences expressed in behavior. 

\subsection{Norms and motivation}

\emph{Norms} are defined as learned utilities of actions, i.e. expected utilities of action. Fundamentally, norms are based on physiological needs, as this is the only way to bootstrap (get starting values for) the values of actions. Utilities can be learned from feedback from agent's own body only. However, the utilities determined by internal rewards may be modified by social interaction: an action with a high internal rewards may cause harm to other agents, who then threat or harm the agent, lowering the utility of the action to take into account the needs of the other agents. Thus, norms of an individual usually partly express utilities of other agents. In a simplified model there is no need to represent the two components separately. They may however be separated when modeling of internal motivational conflicts is required. 

A \emph{standard} is defined as a synonym for norm, though as a term it has a more personal connotation, i.e. internal rewards may dominate over the external rewards. 
\emph{Motivation} equals the expected utility of an action. Motivation and norm are thus synonymous. 

\subsection{The processing loop}

The processing loop of the agent is the following: perceive new events, determine their utilities, update object model, perform the action maximizing utility in the current situation. 
As a new event is perceived, the representation of the causing object is updated to include the utility of the current event. 

The object representation currently being retrieved and updated is defined as being the target of attention. After evaluating all new objects, the object with the highest absolute utility (of all objects in the model) is taken as a target of attention.

\subsection{Object contexts}

If an agent's expected future utility, which it attempts to maximize, is calculated as a sum of utilities of all known objects, it can change only when new events are perceived. However, if it is calculated from conscious objects only, or taking the conscious objects as a starting node and expanding the context from there, keeping low-valued objects unconscious becomes motivated. 
Now e.g. the idea of \emph{repression}  becomes definable. 

Thus, introduction of an \emph{internal object context} enables internal dynamics of the expected future utility. Two kinds of dynamics emerge: first related to new objects, and second related to context switches, which happen during the processing of new events. 

It can also be defined, that agents can \emph{expand} the context. This expansion is conceptualized as an action, which is selected according to the same principle as other actions, i.e. when it is expected to maximize future utility. This may happen e.g. during idle times, i.e. when there are no new events and all pending actions have been performed, or when several actions cannot be prioritized over each other. Expansion or contraction of the context causes context switches and thus potentially changes in expected future utility. 

An especially interesting consequence is that the idea of context expansion during idle times leads to the amount of stress being related to the size of the context (an "emergent" feature). When the agent is overloaded, it context expansion may not take first priority. It "does not have time" to expand the context, i.e. think things thoroughly. Therefore, consciousness of objects' features diminishes; consciousness "becomes shallow". This shallowness includes all object representations, also the self-representation. 

Overloading has also another consequence. New percepts must be evaluated and appropriate actions selected, but there may be no time to perform these actions, which are then queued. The priorities of the queued actions may change when new events are evaluated. Therefore, at each time point a different action has first priority. Actions taking more time than one time unit are started but not finished, since at the next time point some other action is more important. Therefore, the agent perceives that it is "too busy to do anything", a common feature of \emph{burnout}. 

In practice, expansion is done by traversing the object network from the currently prioritized object towards higher utility. For example, an agent has perceived a threatening object and thus expects a negative event in the near future. It targets an affect of fear towards the object. As a result its body state changes to "fear" state. 

One way of conceptualizing action selection would be to think that a list of actions is browsed to see if there is an action that would cancel the threat. 
Another way is to think of the action as a node in the object network. 
Taking the feared object as a starting point, the network is traversed to find a suitable action represented by a node linked with the feared object, the link representing the expected utility of the node. If the node is has the highest utility of all the nodes starting from this object, it is traversed to. If the node is an action, it is performed. It it is another object, the expansion continues. 

As the expected future utility is calculated from the objects in the context, the threat is cancelled when an action with a high enough utility is found, although it may not yet be performed (the utility should be weighted by the probability of succeeding in performing the action). 
This in effect corresponds to a discounting of expected utility. 

Another, probably better, option would be to take the affective state as a starting node. If the agent has previously experienced a state of fear, it has a representation of this state (an object), and actions associated with the state. 

Personality was previously defined as the learned contents of the controlling part of the control system. 
Personality is therefore formed by adding new objects and their associated utilities to the object network.  In the psychoanalytic tradition this is called \emph{internalization} \cite{tahka93}. 

The continuing process of internalizing new, more satisfying functions of the self may be called \emph{progression}. In progression, an agent's focus shifts on the new objects, since the old objects turn out less satisfying in comparison. Correspondingly, if the new functions later turn out to be useless and better ones cannot be found, the agent turns back to the old objects; this may be called \emph{regression}.

\section{Affects, emotions and feelings}
\label{affects}

\subsection{Affect as a bodily process}

Affects are defined here as predefined bodily processes that have certain triggers. When a specific trigger is perceived, a corresponding change in body state follows. 

This change may then be perceived or not. If it is perceived, the content of perception is the process of change. In other words, an affect is perceived when the content of the perception is a representation of  \emph{the body state in transition}, associated with the perception of the trigger. This is essentially the idea of Damasio \cite{damasio03spinoza}.

The triggers are not simple objects, but specific \emph{constellations of object relations}. A certain constellation triggers a certain emotion. For example, fear is triggered when a negative event is expected to happen in the future. There is thus an object relation between the agent and the feared object, in which the object has a negative expected utility. This relation may be seen as an object constellation. In principle the current affect is determined by the whole history of interactions between the agent and the objects, not just the current event, since if e.g. the expected utility was very high in the beginning, a small negative change would not suffice to change the object relation from hope to fear. Alternatively, if an agent knows how to avoid the threat (has an appropriate action), then fear is removed when a representation of the suitable action is retrieved from memory. In such case the agent was expecting to be able to cancel the effects of the expected negative event, and expected utility rises back to a level corresponding to the sum of utilities of the event and the reparative action. 

These differences are however related to triggers only. What makes an \emph{experience} of fear different from an experience of e.g. hope are the perceived differences in bodily reactions associated with these emotions, i.e. a representation of body state associated with one emotion is different from the representation of a representation of another emotion. This is essentially the 'qualia' problem, which in this context would be equal to asking why e.g. fear \emph{feels} like fear, or what gives fear \emph{the quality of fearness}. The solution is that the 'quality' of feeling of e.g. fear is just the specific, unique representation of the body state. There cannot be any additional aspects in the experience; what is experienced (i.e. the target of attention) is simply the representation.

\begin{figure}
  \centering
  \includegraphics[scale=0.48,angle=-90]{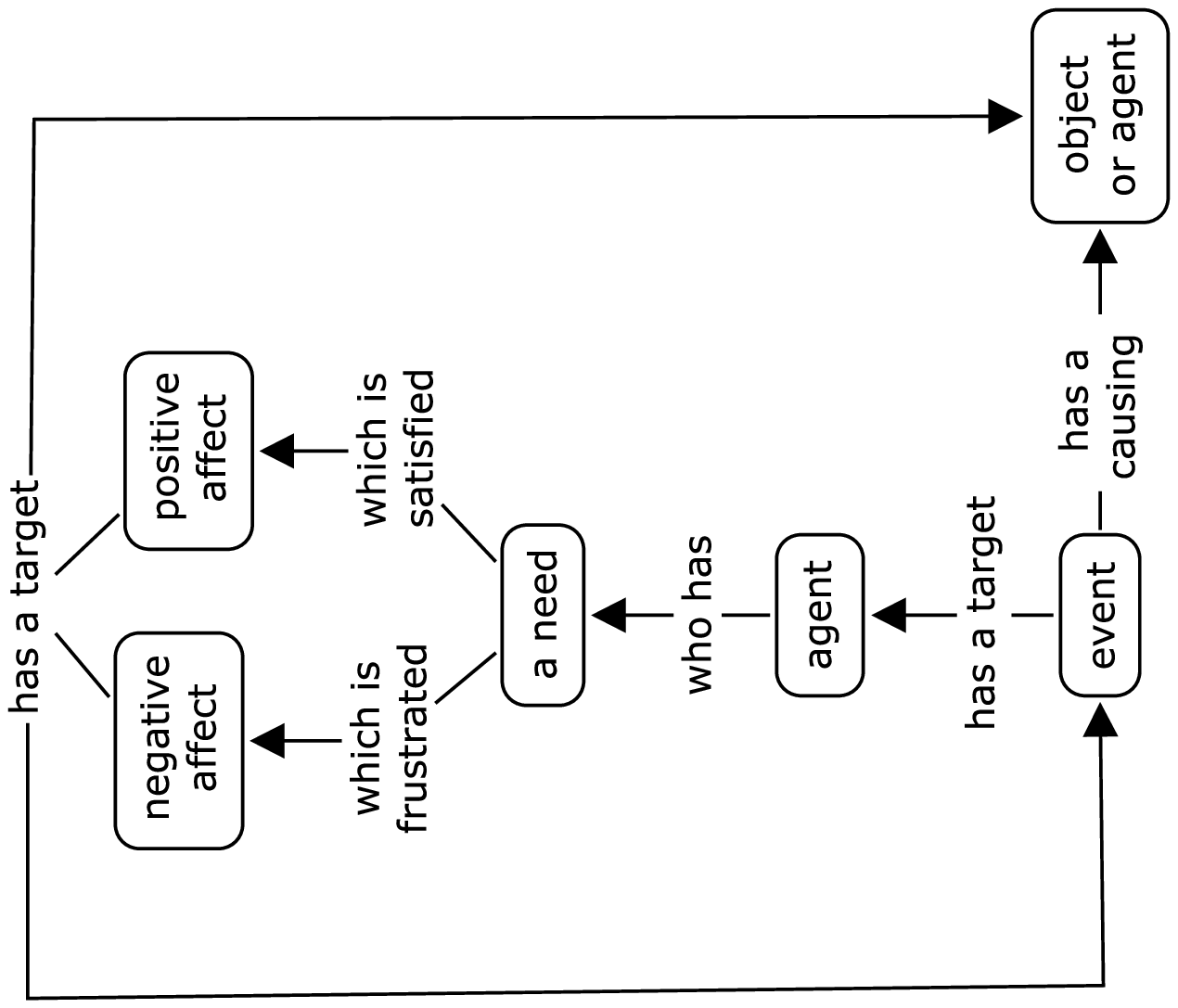}
  \label{fig:eventchart}
  \caption{Relations between event-related concepts.}
\end{figure}

\subsection{Differentiating by levels of consciousness}
\label{consciousness}

Relations between affects, emotions and feelings are defined according to Matthis, who defines affects as a superclass of emotions and feelings \cite{matthis00}. Differentiation is made with respect to levels of consciousness. Emotions are preconscious affects, whereas feelings are conscious affects. There may also be affects that cannot be preconscious or conscious (i.e. cannot be perceived); these are labeled unconscious. 

Now we seem to face the problem of defining consciousness. However, the agent only has to be \emph{conscious of some objects}. E.g. Baars has suggested, that being conscious of an object corresponds to that object being the \emph{target of attention} \cite{baars97}. Let us thus define that conscious contents are the contents that are the target of attention at a given moment. Correspondingly, preconscious are the contents that can be taken at the target of attention, if they become the most important at a given moment. 

When an object is perceived (as a part of an event), an agent searches its internal object model to see if the object is known or unknown. It then attempts to estimate the utility of the event (good or useful, meaningless, bad or harmful) by using the known utility of the object. The internal model of this object is then updated with the utility of the current event. If there is no need to search and no unchecked objects are present, attention is targeted towards the object or action which has the highest absolute value of expected utility. The idea behind this is that utility is maximized by pursuing the highest positive opportunity or dodging the worst threat. If only one goal is present, a higher positive event cancels out a lesser negative event. Multiple goals create more complicated situations, which are not discussed in this article. 

\subsection{Multilayered controlling systems}

For body states in transition and in association with a perceived triggering object constellation to be taken as targets of attention, the controlling system needs an ability to inspect its own structural configurations and their changes in time. Therefore an another layer is needed, that records the states of a lower layer of the controlling system. These records of state change sequences can then be handled as objects and attention can be targeted at them, thus making them preconscious or conscious. 

Unconscious affects are then first-layer affects that cannot be perceived by the second layer. This may be due to e.g. fixed structural limitations in the introspection mechanism. 
Defined this way we can also say that there may be affective agents that are not emotional. In particular, all agents with one-layer controlling system would be affective only. An affective agent can thus be fully unconscious. However, an emotional or a feeling agent needs consciousness. 

\subsection{Differentiating by object constellations}

\begin{figure*}[p]
   \centering
   \includegraphics[scale=1.1,angle=90.0]{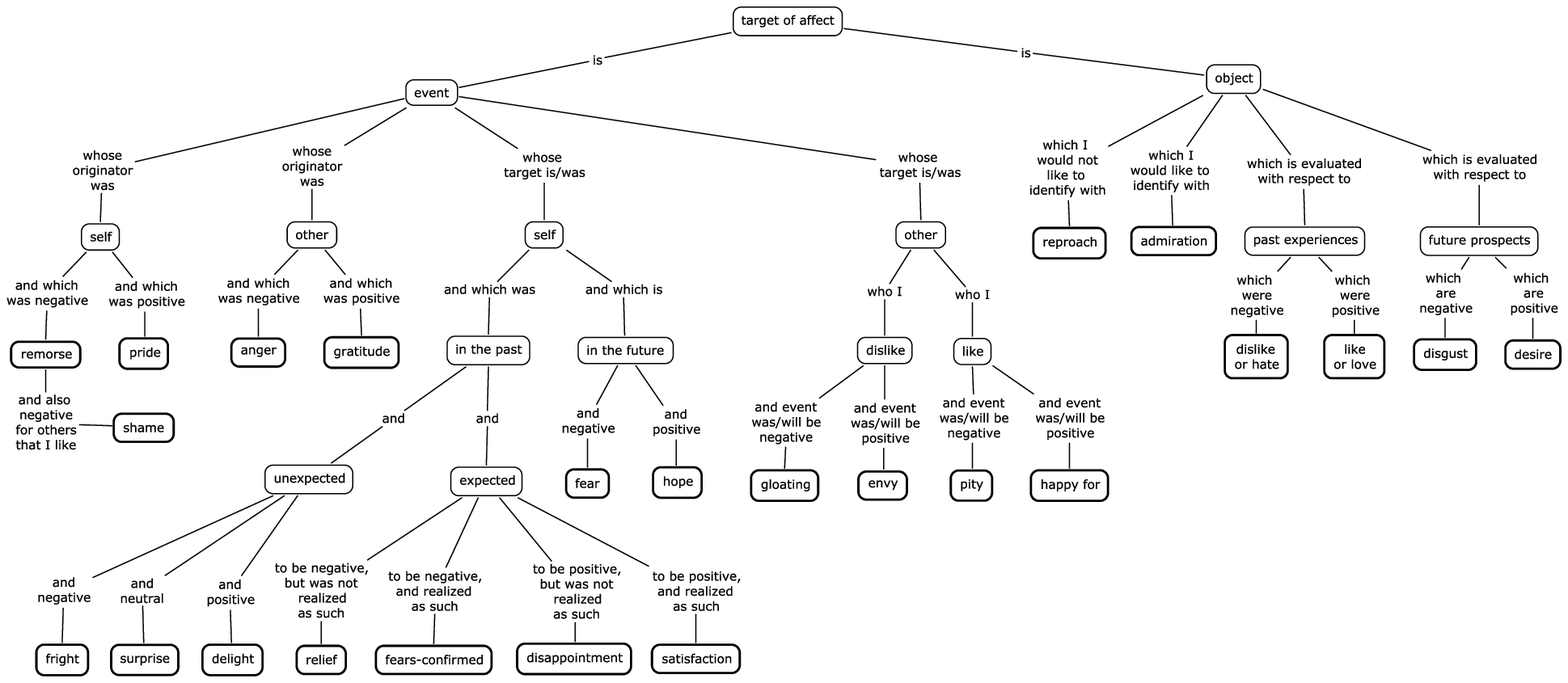}
   \caption{Affects in relation to each other.}
   \label{fig:affectchart}
\end{figure*}

Classification presented here contains mostly the same affects as the OCC model \cite{occ88}, but the classification criteria differ. 
The classification is presented in figure \ref{fig:affectchart}, which may be compared with the classification proposed in the OCC model \cite[p. 19]{occ88}. 

The differentiation criteria are: 
\emph{nature of the target}: whether the target of affect is an event, or an object or agent; 
\emph{time}: whether the event has happened in the past or is expected in the future; 
\emph{expectedness}: whether the object was known or unknown, or whether a past event was expected or unexpected; 
\emph{goal correspondence}: whether the event contributed positively or negatively to agent's goals; 
\emph{self-inflictedness}: whether the event was self-inflicted or caused by others; 
\emph{relation to the target}: whether the target object or agent of the event was liked or disliked. 

A simplified implementation of these criteria can be constructed as follows: agents do not form memories of events as a whole, but only record utilities of causing objects. Future expectations are thus implicit and consist of object utilities only. In other words, agents do not expect specific events, but expect a specific object to have an utility that is the average of the previous events created by it. An object is expected, if a model of it exists, i.e. it has been perceived before as a causing object. Goal correspondence is implicit in the utilities, as agents only have one goal: maximization of the utility. Goal structure and goal derivatives are thus abstracted away in this simplification.

\subsection{Affects related to events}

The first differentiation criteria for event-related affects are:  whether the event was targeted towards self or towards other; and whether the originator of the event was self or other. 

\subsubsection{Events targeted towards self}

\paragraph{Unexpected past events}

Fright is an affect caused by a negative unexpected event. 
Correspondingly, delight is an affect caused by a positive unexpected event.
Surprise is caused by a neutral unexpected event. Whether or not it is an affect is often disputed. If it is associated with e.g. memory-related physiological changes, it would be an affect. Another criteria is, that it is associated with a typical facial expression; in this sense it should be classified as an affect. 

\paragraph{Expected future events}

An expected positive future event causes hope. Correspondingly, an expected negative future event causes fear. 

\paragraph{Expected past events}

Relief is an affect caused by an expected negative event not being realized. 
Disappointment is an affect caused by an expected positive event not being realized. 
Satisfaction is an affect caused by an expected positive event being realized as expected. 
Fears-confirmed is an affect caused by an expected negative event being realized as expected. 

\subsubsection{Events targeted towards others} 

\paragraph{Disliked objects}

Envy is targeted towards a disliked agent that experienced a positive event. 
Gloating is targeted towards a disliked agent that experienced a negative event. 

\paragraph{Liked objects}

Pity is targeted towards a liked agent that experienced a negative event. 
Happy-for is targeted towards a liked agent that experienced a positive event. 

\subsubsection{Self-caused events}

Remorse is targeted torwards a self-originated \emph{action} that caused a negative event to self or someone liked; events positive for disliked objects are considered negative for self. 
Pride is targeted towards a self-originated \emph{action} that caused a positive event to self or a liked object; events negative for disliked objects are considered positive for self. 
Shame is targeted towards \emph{self} when a self-originated action caused a negative event. 

\subsubsection{Events caused by others}

Gratitude is targeted towards an agent that caused a positive event towards self or someone who self depends on (i.e. likes). 
Correspondingly, anger is targeted towards an agent that caused a negative event. 

\subsection{Affects related to agents and objects}

In addition to event-related affects, also the originators and targets of events are targets of affects. 

\subsubsection{Past consequences related affects}

Consequences of events cause the originators of the events to be liked or disliked. 
Like and dislike can be thought of as aggregate terms, taking into account all events caused by an agent. Dislike or hate is targeted towards an agent, who has on average produced more harm than good. Accordingly, like or love is targeted towards an agent, who has produced more good than harm. 
The difference between e.g. like and love is that of magnitude, not of quality; i.e. love is "stronger" liking. A possibly more appropriate interpretation of love as altruism, i.e. as prioritizing needs of others instead of own needs, is currently out of scope of this model. 

\subsubsection{Future prospects related affects}

Future prospects are estimated on the basis of past experiences; therefore they are determined by the past. However, if we set the point of view on the future only, we can differentiate disgust from dislike and like from desire. Desire is an affect caused by a positive future expectation associated to an object. Accordingly, disgust is an affect caused by a negative future expectation. 

\subsubsection{Identification-related affects}

Identification-related affects are currently out of the scope of the computational implementation, as the concept of identification has not been been implemented. 
Agent wants to identify with an object, that has capabilities that would fulfill its needs; in other words, if the object can perform actions that the agent would like to learn. 
Admiration is defined as an affect targeted towards an agent or object that the agent wants to identify with. Accordingly, reproach is targeted towards an object that the agent does not want to identify with. 

\subsubsection{Self-referencing concepts}

Above concepts referred to external objects or relations between objects. Affective concepts can also refer to the object itself: e.g. \emph{mood} refers to the state of the agent itself. Examples of mood are happiness, sadness, depression and mania. A simple definition of happiness could be that the average utility of all events (or events in the context) is above zero (below zero for sadness, respectively). Depression could be defined as a condition where no known objects have a positive utility. 

\subsection{Affects and time}

Often mood is thought of as being somehow qualitatively different from emotions. In this paper, the longer duration of mood is thought to be simply a consequence of the stability of the contents of the object model, which in turn depends on the environment. If the environment does not affect the relevant needs, the affective state does not change.

\section{Demonstration}

A simple browser-based implementation is available at 
\small{\url{http://www.cs.helsinki.fi/u/turkia/emotion/emotioneditor}} . 
In this simulation the user provides events that change the affective states of three agents. 

An example run is as follows. Agent 1 gives agent 2 an utility of 1. Since in the beginning the agents don't have models of each other, this positive event is unexpected, and agent 2 is thus delighted. Also, it now beings to expect a positive utility from agent 1, is begins to like agent 1. In turn, agent 2 gives agent 1 an utility of 1; agent 1 is similarly delighted. 

Now, agent 3 gives agent 1 an utility of zero. Agent 1 is surpised, and it's attitude towards agent 3 is set to neutral. Agent 3 then gives an utility of -1 to agent 2, who is frightened, and begins to dislike agent 3. Although agent 1 is an outsider in this event it reacts to it, since it likes agent 2. Thus, agent 1 targets an affect of pity/compassion towards agent 2 and anger towards agent 3. 

Agent 2 now gives an utility of -2 to agent 3, who is frightened, and begins to dislike agent 2. Agent 2 gloats over the misfortune of agent 3 and feels pride of its own action. Agent 1 feels pity towards 3 and anger at 2 (due to neutral attitude being defined equal to liking). 

Finally, agent 1 gives an utility of 2 to agent 3, who is delighted. Agent 1 feels happy for agent 3 and pride for its own action. Agent 2 feels envy towards the disliked agent 3 and anger towards agent 1. At this point, all agents have expectations of each other. 

Agent 2 now accidentally gives an utility 2 to the disliked agent 3, after which it feels remorse and anger towards self and envy towards 3. Agent 1 feels happy for 3 and gratitude towards 2. 

At this point, agents don't have utilities for themselves. To demonstrate affects related to expected past events, agent 2 gives itself an utility of 2. It is now delighted, likes itself, and expects a similar result in the future. When performing the same event again, agent 2 feels satisfaction and joy. However, now giving itself an utility of 1, it is disappointed and feels remorse. 

As a result of the previous event history, agent 3 expect an utility of 2 and is in a good mood. It does not have expectations of itself. When giving itself an utility of -4, its average expectations change to -2, its expectations towards itself to -4 and its mood to bad. When giving itself an utility of -4 again, its fears are confirmed. When giving itself an utility of -2, it feels relief. 

The event sequence was thus (1,2,1), (2,1,1), (3,1,0), (3,2,-1), (2,3,-2), (1,3,2), (2,3,2), (2,2,2), (2,2,2), (2,2,1), (3,3,-4), (3,3,-4), (3,3,-2), where the first argument of the triple is the causing agent, the second is the target agent, and the third is the utility of the event. As mentioned before, the resulting intersubjective utility expectations form a network of object relations.

\section{Conclusion}

This article presented definitions of affects, that remove the limitations of the OCC model and are easily computationally implementable. They can be used as a starting point in the development of computational psychological, psychiatric, sociological and criminological theories, or in e.g. computer games.

\section*{Acknowledgment}

The author would like to thank Krista Lagus (Helsinki University of Technology), Matti Nyk{\"{a}}nen (University of Kuopio), Ari Rantanen (University of Helsinki) and Timo Honkela (Helsinki University of Technology) for support. 


\vfill

\end{document}